\documentclass{ecai}
\usepackage{graphicx}
\usepackage{latexsym}
\usepackage{inconsolata}
\usepackage{times}  
\usepackage{helvet}  
\usepackage{courier}  
\usepackage{url}  
\usepackage{makecell}
\usepackage{graphicx}  
\usepackage{amsmath}
\usepackage{amsfonts,amssymb}
\usepackage{enumerate}
\usepackage{caption}
\usepackage{epsfig}
\usepackage{threeparttable}
\usepackage{booktabs}
\usepackage{makecell}
\usepackage{array}
\usepackage{multirow}
\usepackage{algorithm, algorithmic}
\renewcommand{\algorithmicrequire}{\textbf{Input:}}
\renewcommand{\algorithmicensure}{\textbf{Output:}}

\renewcommand{\makeorcids}{%
\includegraphics[width=5.2in]{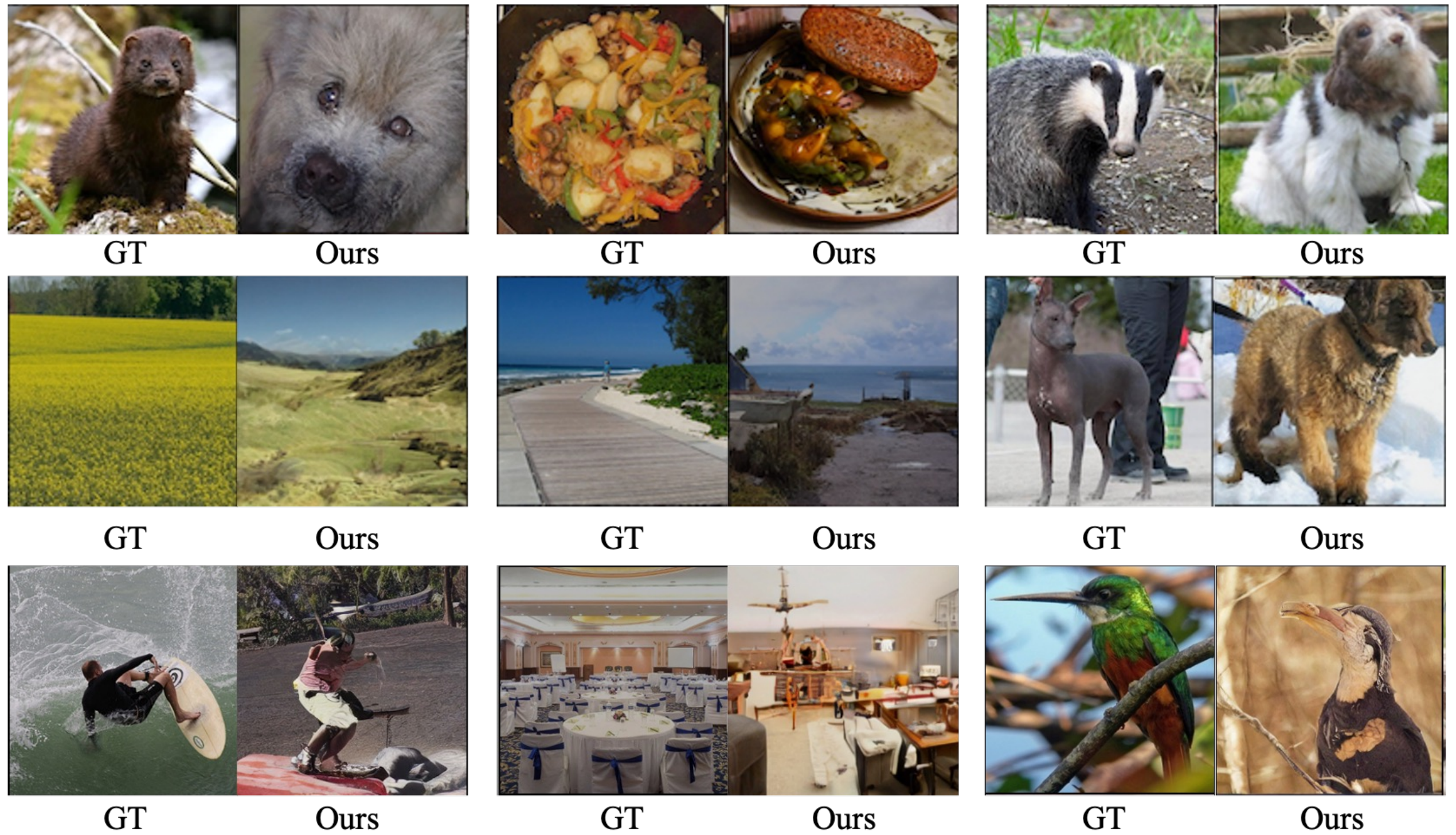}%
\captionof{figure}{The presented ground-truth (GT) image and images decoded from fMRI brain recordings with our method CnD.}
\vskip 3em
}

\begin{document}

\begin{frontmatter}
\title{Decoding Realistic Images from Brain Activity with Contrastive Self-supervision and Latent Diffusion}

\author{Jingyuan Sun\dag\ddag}
\author{Mingxiao Li\dag}
\author{Marie-Francine Moens}

\address{Department of Computer Science, KU Leuven, Belgium}

\address[\ddag]{Corresponding Author (jingyuan.sun@kuleuven.be),${}^\dag$Equal Contribution, }

\begin{abstract}
 Reconstructing visual stimuli from human brain activities provides a promising opportunity to advance our understanding of the brain's visual system and its connection with computer vision models. Although deep generative models have been employed for this task, the challenge of generating high-quality images with accurate semantics persists due to the intricate underlying representations of brain signals and the limited availability of parallel data. In this paper, we propose a two-phase framework named Contrast and Diffuse (CnD) to decode realistic images from functional magnetic resonance imaging (fMRI) recordings. In the first phase, we acquire representations of fMRI data through self-supervised contrastive learning. In the second phase, the encoded fMRI representations condition the diffusion model to reconstruct visual stimulus through our proposed concept-aware conditioning method. Experimental results show that CnD reconstructs highly plausible images on challenging benchmarks. We also provide a quantitative interpretation of the connection between the latent diffusion model (LDM) components and the human brain's visual system. In summary, we present an effective approach for reconstructing visual stimuli based on human brain activity and offer a novel framework to understand the relationship between the diffusion model and the human brain visual system. The code is released at \url{https://github.com/Mingxiao-Li/BrainDecoding}.

\end{abstract}
\end{frontmatter}

\section{Introduction}
Reconstructing visual stimuli from neural imaging data is a promising avenue at the intersection between cognitive neuroscience and machine learning \cite{Horikawa2015GenericDO}. A system that accurately decodes the neural responses to visual input can shed light on the mechanisms of the brain's visual perception and cognition, and help interpret the relation between the human visual system and computer vision models \cite{Ren2021ReconstructingSI,takagi2022high,sun2023ijcai}. Furthermore, it has the potential to form a brain-machine interface that helps patients especially those with motor disabilities to express their thoughts and intentions through brain signals \cite{sun2019towards,sun2020neural}.

Despite its potential, the robust and plausible reconstruction from brain recordings is challenging \cite{Mozafari2020ReconstructingNS}.
The human brain's underlying representations are complex, dynamic, and still largely unknown \cite{quiroga2005invariant,Wang2013AnalysisOW,wang2022fmri}.
Neural responses to visual stimuli are not simply linear mappings of the perceived features but can be considerably influenced by one's knowledge and experience, meaning that different individuals' responses to the same stimulus can diverge significantly \cite{aguirre2016patterns}. Such diversity is further complicated given the biological variability in brain structure. Moreover, the publicly available parallel datasets between brain recordings and  visual stimuli are scarce. 
Datasets scanned by functional magnetic resonance imaging (fMRI) are most frequently used for visual reconstruction tasks, with generally thousands of fMRI-image pairs for each subject available in these datasets.

Researchers have started to address the reconstruction task in recent years with the help of both traditional statistical methods such as ridge regression and deep learning models for example GANs \cite{Fang2021ReconstructingPI,Mozafari2020ReconstructingNS}.
But challenged by the complex pattern and the limited scale of the fMRI-image parallel data, the decoded images by most of the existing methods are not optimal in accuracy and fidelity. To overcome these challenges, we propose that it is first necessary to catch the most informative features in different images and find the common patterns shared among populations over the individual variation. It is secondly important to learn reliable representations efficiently from limited data. 
Contrastive learning naturally meets the first requirement since it aims to group similar samples while separating dissimilar instances. For the second requirement, self-supervised or unsupervised pre-training have been proven to formulate a powerful contextual representation space which effectively supports few-shot learning \cite{Radford2019LanguageMA}. 

Motivated by the above analysis, we propose a two-phase framework based on contrastive self-supervision and latent diffusion models to decode high-quality images from fMRI recordings. In the first phase, we pre-train an fMRI feature learner with a proposed double-contrastive self-supervision loss. In the second phase, inspired by the fact that before visualizing or imaging the appearance of a concept, humans typically first conceive it mentally, we leverage the pre-trained class conditional latent diffusion model and propose the concept aware conditioning, where we utilize a cross-attention module to allow the fMRI feature acquires concept information from the pre-trained concept bank. Experimental results demonstrate that our proposed model can generate high-resolution and semantically accurate images. We also link representations learned by various components and from different denoising stages of the LDM model with the human brain's visual system to interpret their relationship.

In summary, our contributions are three-folded: (1) We propose a contrastive learning and diffusion model based two-phase visual decoding framework, which can generate high-quality and semantic similar images from given fMRI signals. (2) We introduce a systematic way of analyzing the deep generative model from the biological perspective. (3) We demonstrate how the diffusion model incorporates information from regions of the human visual system to reconstruct realistic images.


\section{Related Work}
\subsection{Vision Decoding from fMRI}

In recent years, there has been growing interest in using fMRI data to reconstruct visual experiences, given its potential to provide insights into how the brain encodes and represents visual information. This task has been explored in various contexts, such as explicitly presented visual stimuli \cite{Ren2021ReconstructingSI,Zhang2022DecodingPI}, imagined content \cite{Horikawa2015GenericDO,Horikawa2013NeuralDO} and perceived emotions \cite{Horikawa2019TheNR}.  Early attempts to identify visual images from fMRI mainly used simple statistical models and handcrafted features \cite{Kay2008IdentifyingNI}. Recent studies have employed deep generative models trained on a large number of naturalistic images and hierarchical image features extracted from pre-trained neural networks, and are used either for classification or to reconstruct the original stimulus \cite{Huth2016DecodingTS}. For example, some recent approaches have used regression models to extract latent fMRI representations, which are then used to fine-tune pre-trained generative models like GANs \cite{Mozafari2020ReconstructingNS, Seeliger2017GenerativeAN}. These methods were shown to produce more plausible and semantically meaningful decoded images. Despite such advancements, generating high-resolution images with reliable semantic fidelity remains challenging due to the low signal-to-noise ratio, small sample size and complex patterns associated with fMRI data.\cite{Fang2021ReconstructingPI}. Recent algorithms like diffusion models (DMs) and latent diffusion models (LDMs) show promise in addressing these limitations and generating diverse high-resolution images with high semantic fidelity. 
They have also been recently applied to the scene of visual decoding and deliver better generation results than traditional models \cite{chenzj,takagi2022high}.

\subsection{Diffusion Probabilistic Models}
Diffusion models are first proposed in~\cite{ddpm_ori} and further improved by~\cite{ddpm, improved_ddpm}. A diffusion model contains a forward diffusion process and a backward denoising process. In the forward process, an image diffuses to a normal Gaussian noise by gradually adding Gaussian noise with $T$ steps. During the backward process, an image is recovered from a normal Gaussian noise by several iterations. The diffusion models~\cite{ddpm} originally operate in the pixel space, and though achieving good performance, they consume a massive amount of computing resources during training and inference. Recently, \cite{stable_diffusion} proposed the latent diffusion model by applying a diffusion model in the image latent space learned using a vector quantization regularized autoencoder~\cite{vqvae}. This latent diffusion model not only generates better images but also significantly reduces computation resources. With the integration of a cross-attention mechanism in the UNet model, the latent diffusion model allows applying different controls in image synthesis, such as text control~\cite{photorealistic,diffusionclip,dreambooth} and image controls in different domains~\cite{control,t2i}. 

\begin{figure*}[ht]
\centering
\small
\includegraphics[width=6.3in]{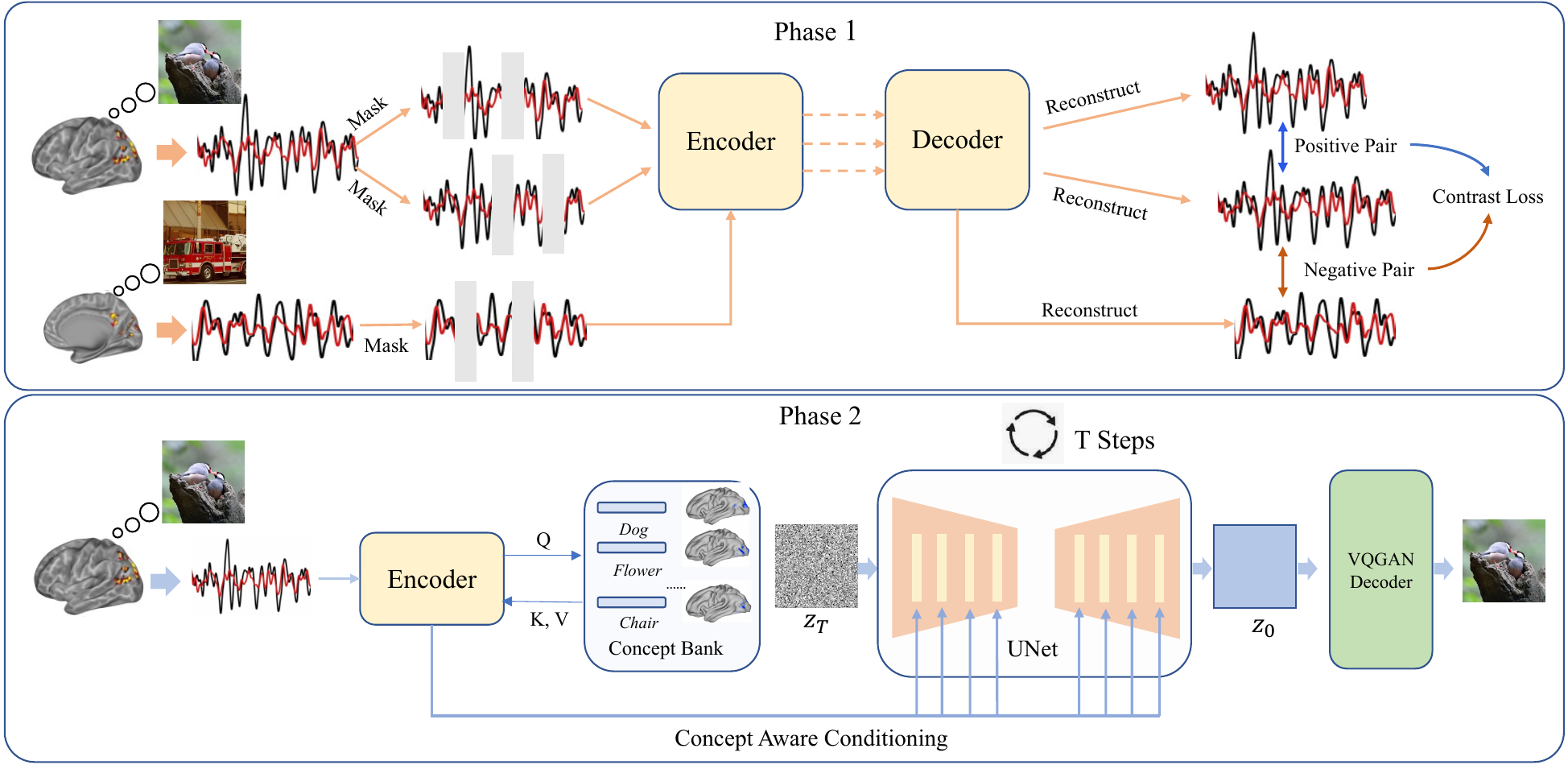}
\caption{Contrast and Diffuse Framework: The framework includes two stages. In the first stage, a ViT\cite{vit})-based encoder is trained with contrastive and reconstruction loss to learn informative fMRI representations; in the second stage, we fix the pre-trained latent diffusion model and introduce the concept-aware conditioning to guide the diffusion model to generate images related to the given fMRI.}
\label{iccv_1}
\end{figure*}

\section{Methods}

\subsection{Preliminary}
\noindent\textbf{Contrastive Learning}\ \
Contrastive learning is a popular technique in unsupervised learning, used to learn useful representations of data. In contrastive learning, the goal is to learn a feature space where similar instances are grouped together, while dissimilar instances are pushed apart. One common way to achieve this is through the use of the InfoNCE loss function~\cite{infonce}, which stands for "InfoMax Contrastive Estimation". The InfoNCE loss is a variant of the contrastive loss that maximizes the mutual information between augmented versions of the same data point while minimizing the mutual information between augmented versions of different data points. The InfoNCE loss has shown to be effective in many applications, such as image recognition and natural language processing.
The InfoNCE loss can be written mathematically as follows:
\begin{equation}
\mathcal{L}_{\text{NCE}}=-\log \frac{\exp(\boldsymbol{z}_{i} \cdot \boldsymbol{z}_{j} / \tau)}{\exp(\boldsymbol{z}_{i} \cdot \boldsymbol{z}_{j} / \tau)+\sum_{k \neq i} \exp(\boldsymbol{z}_{i} \cdot \boldsymbol{z}_{k} / \tau)}
\end{equation}
Here, $\boldsymbol{z}_i$ and $\boldsymbol{z}_j$ are the representations of two positive samples, $\tau$ is a temperature parameter that controls the softness of the distribution, and the sum is taken over all negative samples $\boldsymbol{z}_k$ except for the positive sample $\boldsymbol{z}_j$. The InfoNCE loss has been shown to be effective in various applications, including image classification, object detection, and natural language processing.

\noindent\textbf{Stable Diffusion}\ \ 
Our proposed fMRI-to-image model is based on the recent state-of-the-art latent diffusion model--Stable Diffusion Model~\cite{stable_diffusion} (SD). The SD consists of two components: one vector quantization regularized autoencoder (VQVAE) and a UNet diffusion model. The training of SD has two stages. In the first stage, the VQVAE is trained to map images to latent space; in the second stage, the VQVAE is frozen, and the UNet diffusion model is trained to do denoising in the VQVAE latent space. More specifically, given an image $I$ and the corresponding latent representation denoted as $x_0$, the objective function of training the UNet diffusion model can be formulated as below:
\begin{equation}
    \mathcal{L}_t^{simple} = E_{t,x_0,\epsilon\sim\mathcal{N}(0,1)}[\|\epsilon - \epsilon_{\theta}(z_t,t)\|_2^2]
\end{equation}
where $z_t$ is the noisy image latent representation, and $\epsilon_{\theta}(,)$ is the parameterized UNet model. During inference, given the Gaussian distribution $z_T$, at each step $t$, the SD predicts the noise $\epsilon_t$ and removes it from the latent representation $z_t$. The representation at the last step $z_0$ is seen as a clean image representation without noise and fed into the decoder of VQVAE to generate natural images.

\subsection{Phase 1: Pre-training with Double Contrastive Loss}

We propose a double contrastive autoencoder to learn the fMRI representations.
By saying "Double contrastive", we mean that the model will conduct two times of contrasting when learning to represent an fMRI example.

First, each image $v_i(i\leq n)$ from one batch of $n$ fMRI examples $v$ will go through a random masking function for two independent times. This yields two masked versions of $v_i$ namely $v_i^{m_1}$ and $v_i^{m_2}$. They will form a positive sample pair for the first comparison. $v_i^{m_1}$ and $v_i^{m_2}$ are tokenized into embeddings by a 1D convolutional layer whose stride equals the patch size, and then feeded to the same encoder respectively. The decoder takes each of the encoded latent representations as input and makes predictions $v_i^{dm_1}$ and $v_i^{dm_2}$. Then the first contrastive loss is:\begin{small}
\begin{equation}
\mathcal{L}_{C}=-\log \frac{\exp \left(v_i^{dm_1} \cdot v_i^{dm_2} / \tau\right)}{\exp \left(v_i^{dm_1}  \cdot v_i^{dm_2}  / \tau\right)+\sum_{k \neq i} \exp \left(v_i^{dm_1} \cdot v_{k}^{dm_1} / \tau\right)}
\end{equation}
\end{small}We denote this first contrastive loss as cross-contrastive loss.

Second, every unmasked original image $v_i(i\leq n)$ and its masked image $v_i^{m}$ are also a natural positive sample pair. $v_i^{dm}$ denotes the predicted image that the decoder outputs.
So the second contrastive loss is: \begin{small}\begin{equation}
\mathcal{L}_{S}=-\log \frac{\exp \left(v_i^{dm} \cdot v_i / \tau\right)}{\exp \left(v_i^{dm}  \cdot v_i  / \tau\right)+\sum_{k \neq i} \exp \left(v_i^{dm} \cdot v_{k}^{dm} / \tau\right)}.
\end{equation}
\end{small}The second contrastive loss is denoted as self-contrastive loss.
Optimizing the self-contrastive loss $\mathcal{L}_{S}$ implicitly optimizes the mask-reconstruction loss taken by some previous work \cite{chenzj}. For both $\mathcal{L}_{C}$ and $\mathcal{L}_{S}$, the negative examples are other instances in the same batch. $\mathcal{L}_{C}$ and $\mathcal{L}_{C}$ are optimized jointly as: 
\begin{equation}
    \mathcal{L} = \alpha_{C} \mathcal{L}_{C} +  \alpha_{S} 
 \mathcal{L}_{S}
\end{equation}
where the two $\alpha$s are hyper-parameters controlling the weight of each loss. Here we should note that to train the cross contrasting in Phase 1, one input fMRI image needs to be randomly masked to form two positive samples. Then this leads to a question of whether both of two masked samples will go through the self-contrasting. We name it a duplicate self-contrasting if two masked samples are both passed to calculate the self-contrastive loss. We will discuss the applicatation of the duplicate self-contrasting in the section 5.1.2's ablation so as to be supported by experimental results.

\subsection{Phase 2: Latent Diffusion with Concept Aware Conditioning}

Given the relatively low signal-to-noise ratio (SNR) of functional magnetic resonance imaging (fMRI) and the limited quantity of fMRI-to-image data pairs, it would be difficult to train an fMRI-to-image generation model from scratch. Thus in this phase, we aim to leverage the fMRI to extract image-related knowledge from the pre-trained latent diffusion model. More specifically, we focus on extracting visual knowledge from the pre-trained label-to-image latent diffusion model.  
We then formulate the visual decoding as a conditional image generation task. Motivated by the observation that humans tend to initially conceive a concept in their mind prior to visualizing or imaging its appearance, we propose the concept aware cross-attention and time-step double conditioning consisting of concept learning and condition injecting. During concept learning, we first utilize the pre-trained encoder to obtain the fMRI feature. Subsequently, these features are learned to gather concepts from the concept bank, which is constructed using pre-trained label embeddings of the diffusion model, through cross-attention: 
$\text{CrossAttention(\textit{Q,K,V})} = \text{Softmax}(\frac{QK^T}{\sqrt{d}})V$ with
\begin{equation}
    Q =W_QE(x), K=W_KEmb(c), V= W_VEmb(c)
\end{equation}
where $E$ is the fMRI encoder and $x$ denotes the fMRI, while $Emb(c)$ means the pre-trained embedding of concepts. $W_Q$, $W_K$ and $W_V$ are learnable network parameters. After concept learning, fMRI features are imbued with the concept awareness acquired in the concept learning stage. We then follow previous works~\cite{stable_diffusion,diffusionbeatgan,chenzj} to conduct both cross-attention and time-steps conditioning with the fMRI features. 

\subsection{Linear Decoding Analysis}

The proposed two-phase image reconstruction model CnD constitutes a highly complex non-linear mapping from the fMRI recordings to the visual stimulus. 
After the model is learned, it is non-trivial to understand how information from different regions of the brain visual system contributes to the stimulus reconstruction.
To gain a deeper understanding of the connection between the diffusion model and the brain vision system, we fit linear regression models to directly decode the brain activation patterns with the LDM's UNet Representations.

The UNet consists of an encoder, a middle block and a decoder. We take the UNet encoder as an example to explain the linear decoding analysis procedure.
As shown in Figure 2, the UNet encoder $E$ takes a Gaussian noise and a condition generated by the fMRI encoder as the input.
The hidden states of $E$'s $i$ layer is denoted as $h_E^i$. 
We reduce $h_E^i$'s feature dimension to 300 using principal component analysis (PCA), yielding $\hat{h}_E^i$.
We then fit a ridge regression model to predict the reduced $h_E^i$ from the brain activation patterns $v$ by optimizing the following loss: $|| W v -\hat{h}_E^i||_2^2+\lambda ||W||_1$
The learned regression weights $W$ are projected to the cortical surface, as implemented by previous work in neural decoding \cite{Huth2016DecodingTS}.
These weights implicitly reflect how voxels from different regions of the human visual system contribute to predicting the diffusion model features.

\section{Experimental Setup}

\subsection{fMRI Datasets}

\textbf{HCP}{ }{ } The Human Connectome Project (HCP) is a brain connectome study that has collected and open-sourced neuroimaging as well as behavioral data on 1,200 healthy young adults, aged 22-35. 
HCP provides the currently largest publicly available MRI data on the human brain which is very suitable for pre-training representations of brain activation patterns. In HCP experiments, 
1113 subjects were scanned by a Siemens Skyra Connectom scanner for 3T MR, while 184 subjects were scanned by a Siemens Magnetom scanner for 7T MR.
The 3T dataset where more subjects were scanned will be used in this paper.

\noindent\textbf{GOD}{ }{ } The Generic Object Decoding Dataset constitutes a purpose-built collection of fMRI data intended for fMRI-based decoding. The dataset was acquired through the presentation of images encompassing 200 representative object categories sourced from the ImageNet (2011, fall release) database. A total of 1,200 images, 8 images from each of 150 object categories, were presented for GOD's training session. The test session consisted of 50 images, 1 image from each of the 50 object categories. The test session categories were distinct from those in the training session and were presented in randomized order across runs. Five subjects sbj\_1 to sbj\_5 participated in the fMRI scanning.

\noindent\textbf{BOLD5000}{ }{ } The BOLD5000 dataset was collected and published by a slow event-related human brain fMRI study. There were 5,254 images incorporated in this study with 
4,916 unique ones, which is one of the largest scale publicly available datasets in this field.  This dataset offers the benefit of high diversity, potentially capturing the complexity and variability of natural visual stimuli. It comprises a selection of 1,916 images showing mostly single objects from ImageNet, 2,000 images featuring multiple objects from COCO, and 1,000 images depicting hand-crafted indoor and outdoor scenes from Scenes, all of which are widely used computer vision datasets. 
Four participants CSI1 - CSI4 took part in this study and were scanned by a 3T Siemens Verio MR scanner with a 32-channel phased array head coil. 

\subsection{Implementation Details.}
\noindent\textbf{Phase 1}\ \ \ We pre-train the fMRI encoding in Phase 1 first on the HCP dataset and then tune it on the training set of BOLD5000 or GOD. We use a 24-layer ViT model with a 1D patch embedder as the encoder. The patch size is 16 and the embedding dimension is 1024. For the double contrastive learning, we set the self-contrastive loss weight as 1 and cross contrastive loss weight as 0.5. The mask ratio on fMRI images is 75\%. 

\noindent\textbf{Phase 2}\ \ We fine-tune only the condition module while keeping the weights of the pre-trained diffusion UNet and label embeddings fixed. We fine-tune the diffusion model for $1000$ steps in all experiments and report the results of the best checkpoint. Following previous works~\cite{chenzj}, we generate images of resolution $256 \times 256$. 

We include an ablation study to clarify how different hyper-parameter settings influence reconstruction accuracy in Section 5.1.2. Refer to supplementary materials for more detailed descriptions on implementations and hyper-parameter settings.

\subsection{Baselines and Evaluation Metric}
\noindent\textbf{Baselines}\ \ \ We compare the proposed CnD model with latest published baselines fMRI-ICGAN~\cite{ozcelik2022reconstruction} and Self-supervised AutoEncoder (SS-AE)~\cite{gaziv2022self}. The first baseline was built upon Instance-Conditioned GAN, while the second one relied on cycle consistency and perceptual losses to reconstruct images from fMRI brain recordings. These are typical methodologies of visual reconstruction. We also notice that a model named DC-LDM~\cite{chenzj} delivers impressive performance on the GOD dataset. But DC-LDM requires tuning on the test set fMRI to achieve its optimal task performance, which is not the evaluation setting taken by our baselines and related previous work. To ensure a fair evaluation, we will not include DC-LDM as a baseline to be compared.

\noindent\textbf{Evaluation metric}\ \ \  In visual decoding, a greater emphasis is placed on semantic consistency. Accordingly, we assess the semantic accuracy of our outcomes using the $n$-way top-1 accuracy metrics, consistent with prior literature~\cite{gaziv2022self}. Specifically, the pre-trained ImageNet1K classifier~\cite{vit} is taken as the semantic correctness evaluator. During the evaluation, both generated image and the corresponding ground truth image are sent to the classifier. The semantic correctness is then defined as if the top-$k$ classification in $n$ randomly selected classes matches the ground-truth classification. For more details, we refer readers to~\cite{chenzj}.

\begin{figure}[h]
\centering
\includegraphics[width=3.3in]{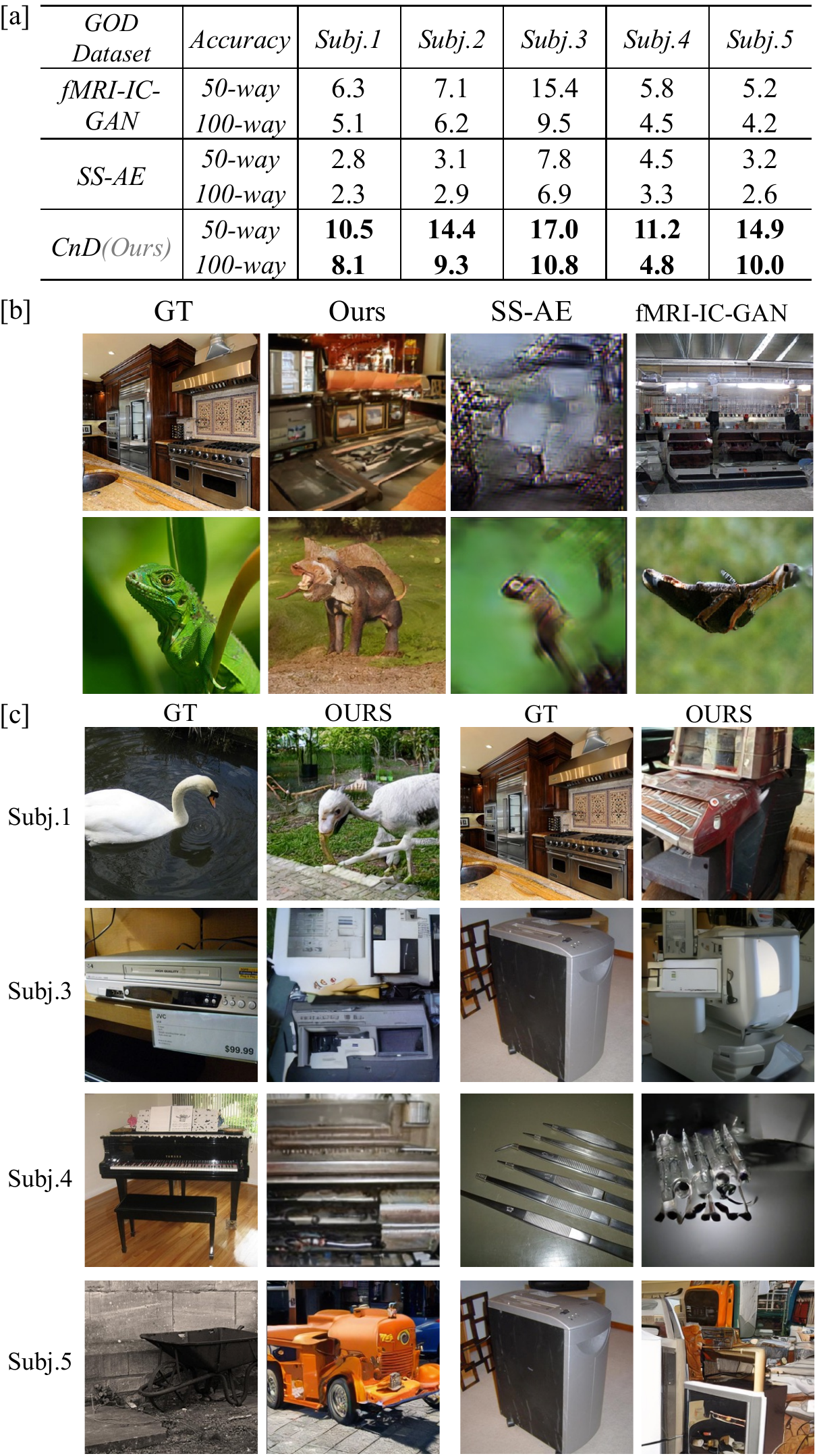}
\caption{[a] Quantitative performance comparisons of our model with other two models on the GOD
test set.  [b] Ground truth images and images generated using different models. Samples are randomly selected from the GOD subject 2 test set. [c] Images generated by our model. Samples are randomly selected from the GOD test set for each subject except subject 2. }
\end{figure}

\section{Results}
\subsection{Image Reconstruction}
\subsubsection{Reconstruction Results Evaluation}
We evaluate our model on both GOD and BOLD5000 datasets, and present their results in Figure 3 and Figure 4.
The results in Table[a] of Figure 3 indicate substantial variability in the performance of all models across diverse subjects.
Our model surpasses the previous fMRI-ICGAN and SS-AE models by a large margin across all subjects and evaluation metrics in GOD dataset. For example, our model achieves around $7.7$ and $3.7$ higher 50-way accuracy than SS-AE and fMRI-ICGAN, respectively. 
To investigate the quality of images generated by different models, we randomly select 2 samples from the GOD Subject 2's test set and present the generated images of our model and baselines in Figure 3[b]. We also show the generated results of our method from the other 4 subjects in Figure 3[c]. It is clear that  our model can generate high-resolution and semantically similar images, while images generated by SS-AE and fMRI-IC-GAN are vague and has a low amount of semantic information.
In Figure 4, we further present generated images of our model on all 4 BOLD5000 subjects. These samples are randomly selected from the test set. Our model can correctly generate a high-resolution image with similar semantic meaning to the ground truth image.

\begin{figure}[h]
\centering
\includegraphics[width=3.2in]{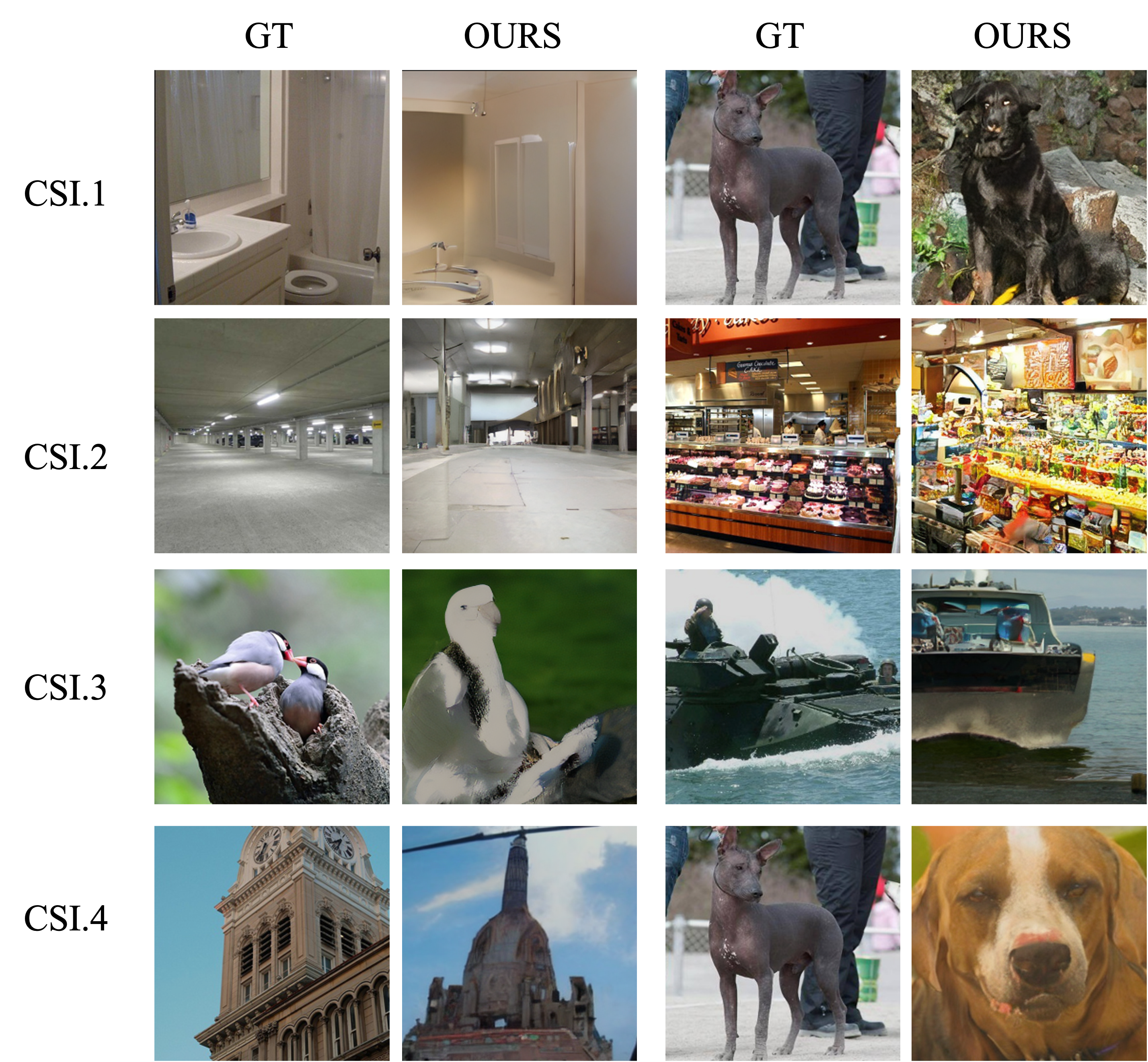}
\caption{
Images generated by our model. Samples are randomly selected from the BOLD test set for each subject.}
\end{figure}

\subsubsection{Ablation Study}

In this section, we conduct an ablation study to clarify the contribution of CnD's modules and how different hyper-parameter settings influence reconstruction accuracy. Without loss of generality, we conduct the experiments on GOD Subj.1.  The results are presented in Table 1.

\begin{figure}[ht]
\centering
\small
\includegraphics[width=3.3in]{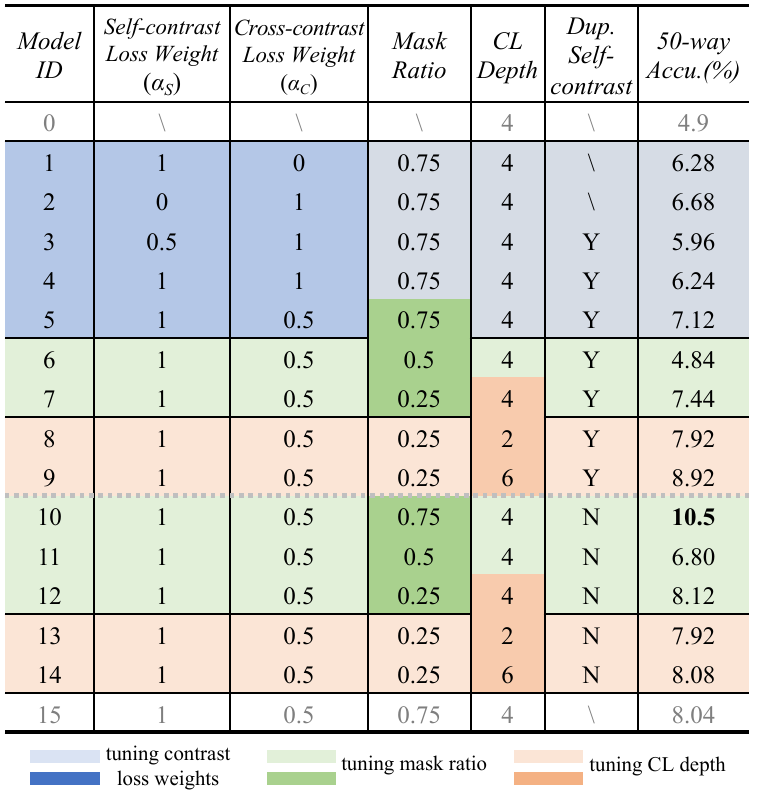}
\caption*{\textbf{Table 1}: The results of ablation study. Cells with the same shade target one same type of hyper-paramter, in which cells with darker shape highlight the different settings of this hyper-parameter. The legend at the bottom explains the meaning of different colors. CL is the abbreviation of the "concept learning", while Dup. means "duplicate". }
\label{tab2}
\end{figure}

\noindent\textbf{Contrastive Loss Weights}\ \ \ The optimization of the double contrastive loss is the central goal of Phase 1. So we begin by investigating the effects of tuning self and cross-contrastive loss weights $\alpha_{S}$ and 
 $\alpha_{C}$. 
 We first evaluate the models with duplicate self-contrasting. 
 Cells in Table 1 with blue shade record the results of tuning $\alpha_{S}$ and $\alpha_{C}$. In model 1 and 2 we set either $\alpha_{S}$ or $\alpha_{C}$ to be 0, which means optimising the self-contrastive or cross-contrastive loss solely. In model 3-5 we then turn on both the contrastive loss and evaluate their combinations. We find that model 5 with $\alpha_{S}=1, \alpha_{C}=0.5$ ranks the top in reconstructing accuracy.
 We also find that using a single contrastive loss leads to better results than combining them together in some models. This can be caused by the entangling of self and cross-contrastive losses  with duplicate self-contrasting. Because when we turn off the duplicate self-contrasting in model 10, it leads to significant improvements over model 5. Model 5 and model 10 only differ in applying or not the duplicate self-contrasting.

\noindent\textbf{Mask Ratio}\ \ \ The double contrastive learning in Phase 1 is built upon the masked ViT. So the mask ratio in the input fMRI data is an important hyper-parameter that may largely influence the model's reconstruction performance. Cells in Table 1 with green shades report the results of tuning the mask ratio. We evaluate respectively with mask ratio 0.25, 0.5 and 0.75 based on the optimal loss weight setting found in last subsection. We find that model 10 with the mask ratio of 0.75 and without duplicate self-contrast produces the best reconstruction accuracy. 

\begin{figure}[h]
\centering
\small
\includegraphics[width=3.3in]{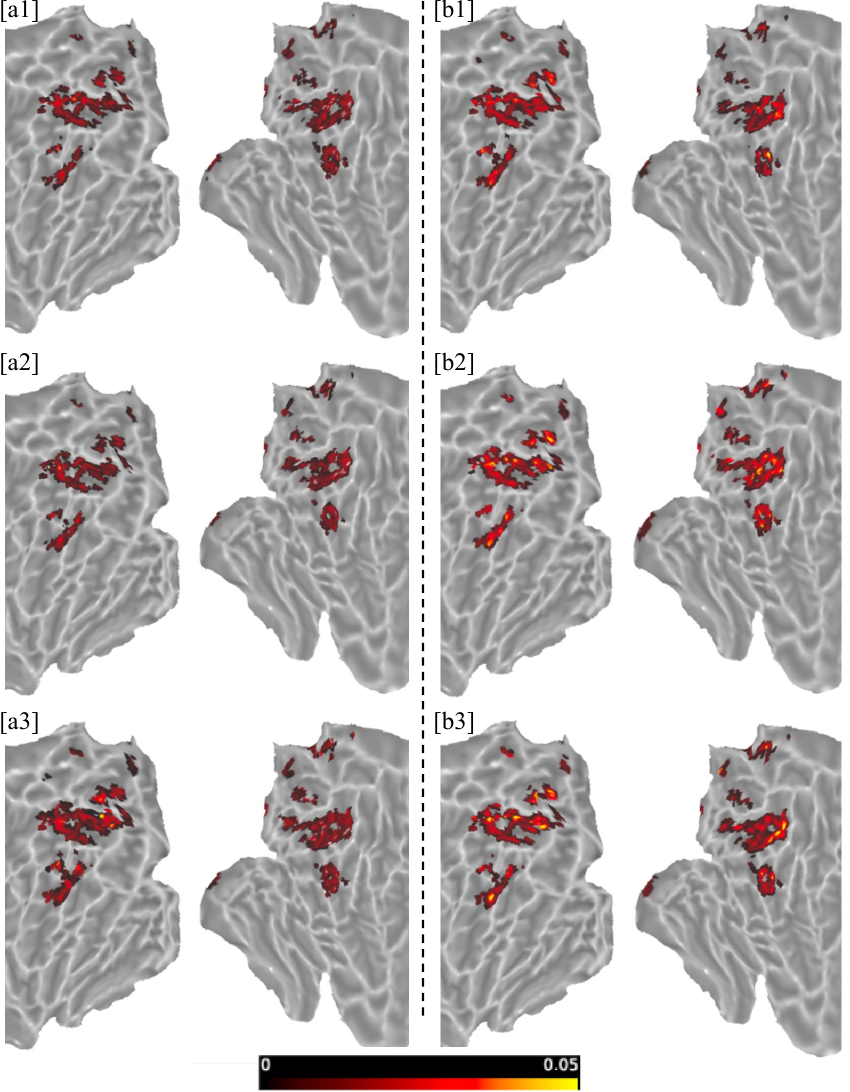}
\caption{Regression weights of decoding different LDM representations from brain activation patterns. Figure a and b respectively depict the weights learned in predicting the UNet encoder and decoder block. Sub-figure 1, 2 and 3 in a and b denote predicting the representation from 4, 9 and 14 layer of encoder (a) and decoder (b). The weights are projected onto the cropped flat cortical surface for more straightforward demonstration. }
\label{iccv_4}
\end{figure}

\noindent\textbf{Concept-Learning Depth}\ \ \ In Phase 2, we propose latent diffusion with concept aware conditioning to achieve image reconstruction. We conduct cross-attention among the encoded fMRI representations and concepts. The number of cross-attention layers can thus be critical to performance of the concept learning (CL) module. We use CL depth to abbreviate the number of cross-attention layers in CL. 
The orange shaded cells in Table 1 demonstrate the effects of tuning CL depth. The results reflect that a middle size of CL depth, that is, 4 layers, is a more optimal setting for our model that does not conduct duplicate self-contrast.

\subsection{Linear Decoding Analysis}
\subsubsection{Decoding with UNet Representations}

We first predict the UNet encoder and decoder representations. Without the loss of generality, we uniformly sample from middle layers and take the 4, 9 and $14^{th}$ layer. We average the representations from each layer and fit the regression models from brain activities. 
The learned weights of the regression model then reflect how much the voxels of different cortical areas contribute to predict the LDM representations. A full denoising process takes a total of 250 steps. Taking in the possible variances of different denoising stages, for each layer we average the weights when predicting the 0, 50, 150 and 250 steps' representation. Figure 5 shows the results.

We find there are both overlaps and discrepancies between the region of interest (ROIs) of which voxels contribute more to predict the LDM representations. 
The Medial IntraParietal Area (MIP) of both hemispheres contain voxels that contribute to predict layers of encoder and decoder of the LDM-UNet.
Studies have shown that the MIP contains neurons that link visual information about object properties, such as shape, size, and orientation, to the motor programs required to interact with those objects.
Voxels that contribute more to predict the UNet encoder representations show similar cortical distribution pattern among the three layers, as displayed in Figure 5[a1-3]. 
But for the decoder,the distribution patterns change when the target layer goes deeper, as shown Figure 5[b1-3]. 
For example, part of the third visual cortex (V3A)  and  posterior half of inferior parietal cortex (PGi) voxels tend to contribute more to predict the representations of deeper decoder layers. 

\subsubsection{Decoding across Diffusion Steps}

We next study how the LDM representations decode the brain activities with the iterative denoising process going. 
A full denoising process takes a total of 250 steps. We acquire the representations produced by the UNet encoder and decoder at 0, 50, 150, and 250 steps. We train regression models to predict these representations from brain activities. Taking in the possible variances caused by layer differences, for each denosing stage we average the weights of predicting the 4, 9 and  $14^{th}$ layer's representation.

Regression weights for predicting encoder and decoder representations are depicted in Figure 6. We observe that with the timestep going forward, the number of voxels that contribute to predict the encoder features increases. But the number of voxels that contribute to predict the encoder features decreases. For example, we show that at the earlier stage of the denoising process,  V3B voxels highly contribute to predict the decoder. But when the denoising process is approaching the end, V3B voxels contribute less to predict the decoder representations. These findings indicate that during denoising the encoder and decoder of diffusion model may focus on capturing different levels of visual information. 

\begin{figure}[ht]
\centering
\small
\includegraphics[width=3.2in]{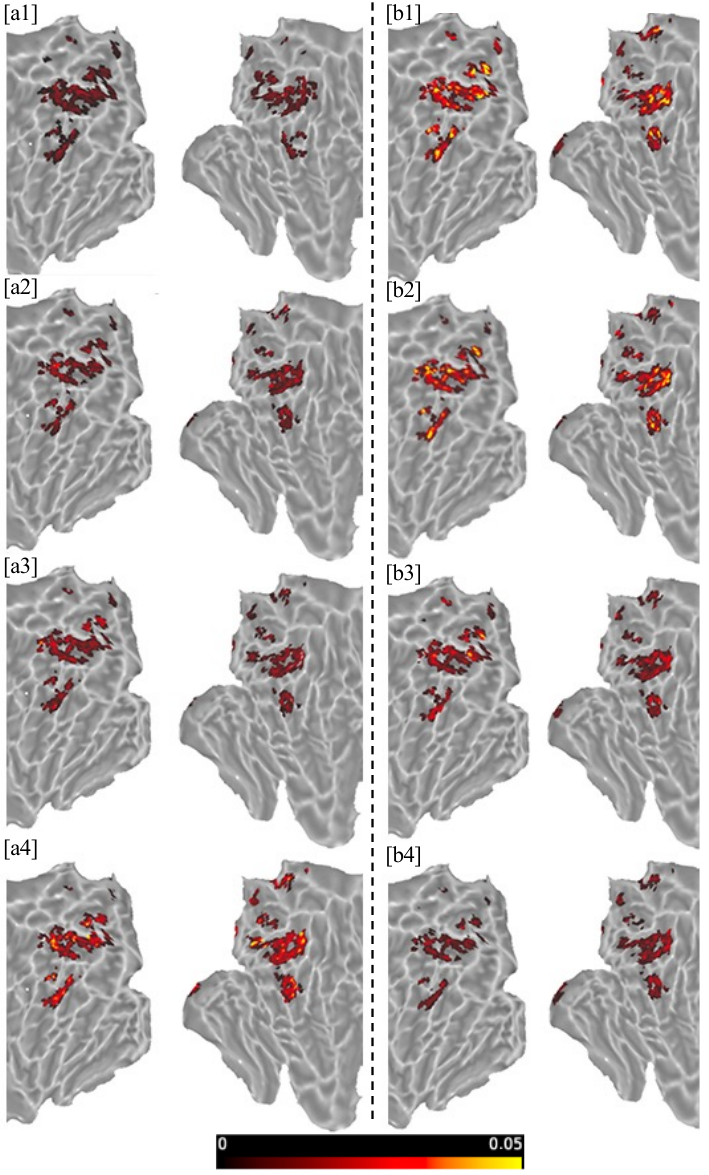}
\caption{Regression weights of predicting UNet encoder (a) and decoder (b) representations from brain activation patterns along the denoising process. The timestep of the denoising process is 0 (a1/b1), 50 (a2/b2), 150 (a3/b3) and 250 (a4/b4) from top to bottom. 
}
\label{iccv_3}
\end{figure}

\section{Discussion}
Our experiment results show that with the help of the LDM we can, to some extent, recover perceived visual information from the human brain activities. Figure 7 presents more examples generated using our model. However, our analysis reveals that the model exhibits a bias towards generating certain categories with greater frequency than others, for example, the model is more likely to generate dogs than elephants when the ground truth is animals. We argue that the potential cause of this phenomenon could be attributed to the bias in the training data of the latent diffusion model. Another limitation of our model is that, though the model can capture high level semantics of the image, sometimes details  are missing in the generated image. For example, in Figure 7 row 2, the ground truth is airplane, for all subjects the model captures the high-level semantic meaning, that is generating something that can fly, but fails to generate airplanes. Different from general image generation focusing on generation diversity, visual decoding relies more on generation consistency, which requires less bias during generation. Thus exploring how to reduce the impact of data bias when generating images from fMRI would hold significant value and academic interest.

\begin{figure}[h]
\centering
\includegraphics[width=3.3in]{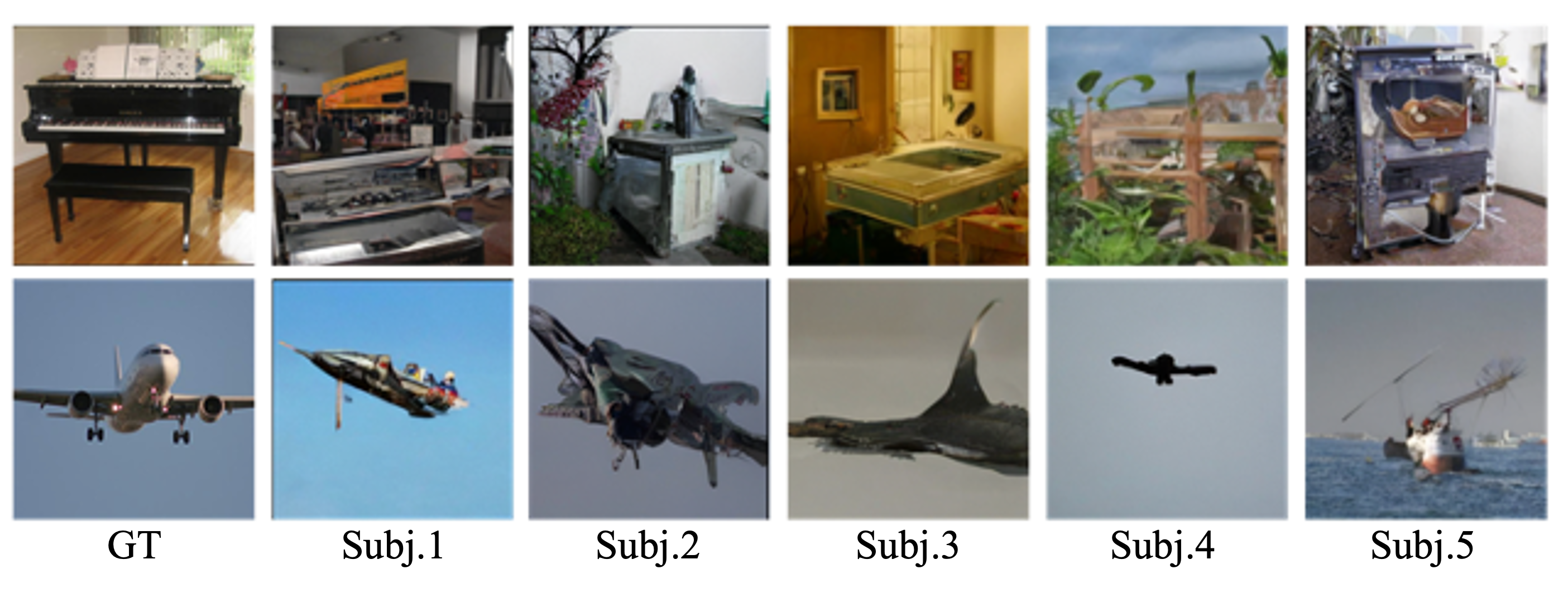}
\caption{Images generated by our model using fMRI signals from different subjects.  Samples are randomly selected from GOD test set.}
\end{figure}

Trough studying the connection bettwen LDM and brain representations, we find overlaps and discrepancies in the region of interest (ROIs) where voxels contribute to predict the LDM representations. The MIP in both hemispheres contain voxels that contribute to the encoder and decoder of the LDM-UNet. The MIP has neurons that link visual information about object properties to motor programs needed to interact with them.  The distributions of voxels that contribute to predicting different layers of encoder vary less than decoder representations. The number of voxels that highly contribute to predict the encoder features increases along the denoising process. But the number of voxels that highly contribute to predict the decoder features decreases. Part of the third visual cortex (V3A) voxels contributes to predicting the decoder representations but not the encoder representations.

\section{Conclusion}
In this work, we first propose a visual decoding model consisting of two stages including contrastive pre-training and concept-aware conditional fine tuning. Experimental reusults illustrate that our model can generate high-quality images from fMRI features. We then further conduct extensive experiments and analyses to understand the information processing  within the pre-trained latent diffusion model by examining the connections between the hidden representations of diffusion UNet and some specific regions in the brain. 

\section*{Acknowledgements}
This work is funded by the CALCULUS project (European Research Council Advanced Grant H2020-ERC-2017-ADG 788506).
This work is also supported by FLAIR project (Flanders AI Impuls Programme).

\bibliography{ecai}

\section{Appendix}
\subsection{Evaluation Metrics}
We use the common N-trial, n-way top-1 semantic classification as the main evaluation metrics. This evaluation method is summarized as in Algorithm below:

\begin{algorithm}[h]
\renewcommand{\algorithmicrequire}{\textbf{Input:}}
\renewcommand{\algorithmicensure}{\textbf{Output:}}
\caption{Iterative Reasoning Module}
\begin{algorithmic}
\REQUIRE ~~\\ pre-trained image classifier $F$, generated image $x$, corresponding ground truth (GT) image $\hat{x}$
\ENSURE ~~\\ success rate $sr \in [0,1]$
\FOR{$trail=1$ to $N$}
\STATE $y_g = F(x_g) $ get the prediction of GT image
\STATE $pred = F(x)$ get the output probabilities of generated image
\STATE $p = \{p_{_g},p_{y_1},...,p_{y_{n-1}}\}$ generate probabilities set contains $n-1$ randomly selected from $pred$ and $y_g$
\STATE Success if $\mathop{\arg\min}\limits_{y} = y_g$
\ENDFOR
\RETURN number of success / N
\end{algorithmic}
\end{algorithm}

\subsection{Implementation Details}

In the Phase 1, we train the masked ViT-based fMRI encoder with contrastive loss.  We employed an asymmetric architecture for the fMRI encoder, in which the decoder is considerably smaller with 8 layers than the encoder with 24 layers. We divided fMRI voxels into patches and transformed them into embeddings using a one-dimensional convolutional layer with a patch size stride. We used a larger embedding to patch size ratio, specifically a patch size of 16 and embedding dimension of 1024 for our model. Our design choice expands the representation dimension of fMRI data, which increases the information capacity of the fMRI representations.  

In the Phase 1, the fMRI encoder is trained by optimized the double contrastive loss where a larger batch size is appreciated. So we set the batch size to be 250 and train for 500 epochs. We train with 20-epoch warming up and max learning rate 2.5e-4. We optimze with AdamW and weight decay 0.05. The initial learning rate is  To address the data-hungry nature of models like the Vision Transformer (ViT), we used random sparsification (RS) as a form of data augmentation, randomly selecting and setting 20\% of voxels in each fMRI to zero.

In the Phase 2,  LDM finetuning stage, we finetune model in all experiments for around 500 epochs with a batch size of $8$. We use AdamW optimizer and the learning rate is $5.3e^{-5}$.
\end{document}